\documentclass[11pt,a4paper]{article}
\usepackage[hyperref]{acl2018}
\usepackage{times}
\usepackage{latexsym}
\usepackage[pdftex]{graphicx}
\usepackage [autostyle, english = american]{csquotes}
\MakeOuterQuote{"}

\usepackage{url}

\aclfinalcopy 

\title{Continuous Learning in a Hierarchical Multiscale Neural Network}

\author{Thomas Wolf, Julien Chaumond \& Clement Delangue \\
Hugging Face Inc.\\
81 Prospect St.\\
Brooklyn, New York 11201, USA\\
{\tt \{thomas,julien,clement\}@huggingface.co} \\}

\begin{document}
\maketitle
\begin{abstract}
We reformulate the problem of encoding a multi-scale representation of a sequence  in a language model by casting it in a continuous learning framework. We propose a hierarchical multi-scale language model in which short time-scale dependencies are encoded in the hidden state of a lower-level recurrent neural network while longer time-scale dependencies are encoded in the dynamic of the lower-level network by having a meta-learner update the weights of the lower-level neural network in an online meta-learning fashion. We use elastic weights consolidation as a higher-level to prevent catastrophic forgetting in our continuous learning framework. 
\end{abstract}

\section{Introduction}

Language models are a major class of natural language processing (NLP) models whose development has lead to major progress in many areas like translation, speech recognition or summarization \citep{schwenk_continuous_2012,arisoy_deep_2012,rush_neural_2015,nallapati_abstractive_2016}. Recently, the task of language modeling has been shown to be an adequate proxy for learning unsupervised representations of high-quality in tasks like text classification \citep{howard_fine-tuned_2018}, sentiment detection \citep{radford_learning_2017} or word vector learning \citep{peters_deep_2018}.

More generally, language modeling is an example of online/sequential prediction task, in which a model tries to predict the next observation given a sequence of past observations. The development of better models for sequential prediction is believed to be beneficial for a wide range of applications like model-based planning or reinforcement learning as these models have to encode some form of memory or causal model of the world to accurately predict a future event given past events.

One of the main issues limiting the performance of language models (LMs) is the problem of capturing long-term dependencies within a sequence.

Neural network based language models \citep{hochreiter_long_1997, cho_learning_2014} learn to implicitly store dependencies in a vector of hidden activities \citep{mikolov_recurrent_2010}. They can be extended by attention mechanisms, memories or caches \citep{bahdanau_neural_2014,tran_recurrent_2016,graves_neural_2014} to capture long-range connections more explicitly. Unfortunately, the very local context is often so highly informative that LMs typically end up using their memories mostly to store short term context \citep{daniluk_frustratingly_2016}.

In this work, we study the possibility of combining short-term representations, stored in neural activations (hidden state), with medium-term representations encoded in a set of dynamical weights of the language model. Our work extends a series of recent experiments on networks with dynamically evolving weights \citep{ba_using_2016, ha_hypernetworks_2016,krause_dynamic_2017,moniz_nested_2018} which show improvements in sequential prediction tasks. We build upon these works by formulating the task as a hierarchical online meta-learning task as detailed below.

The motivation behind this work stems from two observations.

On the one hand, there is evidence from a physiological point of view that time-coherent processes like working memory can involve differing mechanisms at differing time-scales. Biological neural activations typically have a 10~ms coherence timescale, while short-term synaptic plasticity can temporarily modulate the dynamic of the neural network it-self on timescales of 100~ms to minutes. Longer time-scales (a few minutes to several hours) see long-term learning kicks in with permanent modifications to neural excitability \citep{tsodyks_neural_1998,abbott_synaptic_2004,barak_persistent_2007,ba_using_2016}. Interestingly, these psychological observations are paralleled, on the computational side, by a series of recent works on recurrent networks with dynamically evolving weights that show benefits from dynamically updating the weights of a network during a sequential task \citep{ba_using_2016, ha_hypernetworks_2016,krause_dynamic_2017,moniz_nested_2018}.

In parallel to that, it has also been shown that temporal data with multiple time-scales dependencies can naturally be encoded in a hierarchical representation where higher-level features are changing slowly to store long time-scale dependencies and lower-level features are changing faster to encode short time-scale dependencies and local timing \citep{schmidhuber_learning_1992,el_hihi_hierarchical_1995,koutnik_clockwork_2014, chung_hierarchical_2016}.

As a consequence, we would like our model to encode information in a multi-scale hierarchical representation where
\begin{enumerate}
\item \textit{short time-scale dependencies} can be encoded in fast-updated neural activations (hidden state),
\item \textit{medium time-scale dependencies} can be encoded in the dynamic of the network by using dynamic weights updated more slowly, and
\item \textit{a long time-scale memory} can be encoded in a static set of parameters of the model.
\end{enumerate}

In the present work, we take as dynamic weights the full set of weights of a RNN language model (usually word embeddings plus recurrent, input and output weights of each recurrent layer).

\section{Dynamical Language Modeling}
Given a sequence of $T$ discrete symbols $S=(w_1, w_2, \dots, w_T)$, the language modeling task consists in assigning a probability to the sequence $P(S)=p(w_1, \dots, w_T)$ which can be written, using the chain-rule, as
\begin{equation}
P(S\mid\theta) = \prod_{t=1}^T P(w_t \mid w_{t-1}, \dots, w_0,\theta)P(w_0\mid\theta).
\end{equation}
where $\theta$ is a set of parameters of the language model.

In the case of a neural-network-based language model, the conditional probability $P(w_t \mid w_{t-1}, \dots, w_0,\theta)$ is typically parametrized using an autoregressive neural network as
\begin{equation}
P(w_t \mid w_{t-1}, \dots, w_0,\theta) = f_\theta(w_{t-1}, \dots, w_0)
\end{equation}
where $\theta$ are the parameters of the neural network.

In a dynamical language modeling framework, the parameters $\theta$ of the language model are not tied over the sequence $S$ but are allowed to evolve. Thus, prior to computing the probability of a future token $w_t$, a set of parameters $\theta_t$ is estimated from the past parameters and tokens as $\theta_{t} = \arg\!\max\limits_{\theta}P(\theta\mid w_{t-1}, \dots, w_0,\theta_{t-1}\dots\theta_{0})$ and the updated parameters $\theta_{t}$ are used to compute the probability of the next token $w_t$.

In our hierarchical neural network language model, the updated parameters $\theta_{t}$ are estimated by a higher level neural network $g$ parametrized by a set of (static) parameters $\phi$:
\begin{equation}
\theta_{t} = g_\phi(w_{t-1}, \dots, w_0,\theta_{t-1}\dots\theta_{0})
\end{equation}

\subsection{Online meta-learning formulation}
\begin{figure*}
\begin{center}
\includegraphics[width=0.8\linewidth]{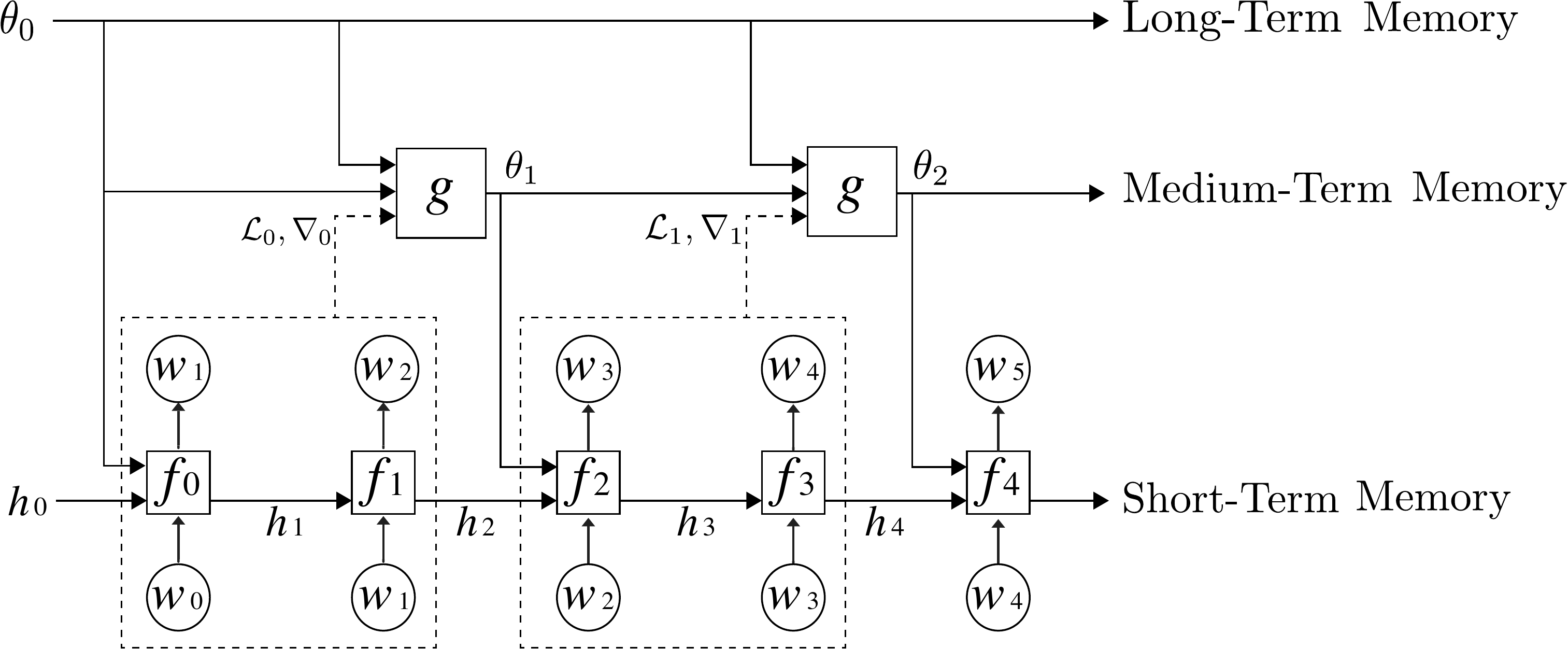}
\end{center}
\caption{A diagram of the Dynamical Language Model. The lower-level neural network $f$ (short-term memory) is a conventional word-level language model where $w_0, \dots, w_5$ are words tokens. The medium-level language model $g$ is a feed-forward or recurrent neural network  while the higher-level memory is formed by a static set of consolidated pre-trained weights (see text).}
\label{fig:fig1}
\end{figure*}

The function computed by the higher level network $g$, estimating $\theta_{t}$ from an history of parameters $\theta_{<t}$ and data points $w_{<t}$, can be seen as an online meta-learning task in which a high-level meta-learner network is trained to update the weights of a low-level network from the loss of the low-level network on a previous batch of data. 

Such a meta-learner can be trained \citep{andrychowicz_learning_2016} to reduce the loss of the low-level network with the idea that it will generalize a gradient descent rule
\begin{equation}
\theta_{t} = \theta_{t-1} - \alpha_t\nabla_{\theta_{t-1}}\mathcal{L}_t
\end{equation}
where $\alpha_t$ is a learning rate at time $t$ and $\nabla_{\theta_{t-1}}\mathcal{L}_t^{LM}$ is the gradient of the loss $\mathcal{L}_t^{LM}$ of the language model on the $t$-th dataset with respect to previous parameters $\theta_{t-1}$.

Ravi and Larochelle (\citeyear{ravi_optimization_2016}) made the observation that such a gradient descent rule bears similarities with the update rule for LSTM cell-states 
\begin{equation}
c_{t} = f_t\odot c_{t-1} +i_t\odot\tilde{c}_t
\end{equation}
when $c_{t}\to\theta_{t}$, $i_t\to\alpha_t$ and $\tilde{c}_t\to-\nabla_{\theta_{t-1}}\mathcal{L}_t$

We extend this analogy to the case of a multi-scale hierarchical recurrent model illustrated on figure~\ref{fig:fig1} and composed of:
\begin{enumerate}
\item \textit{Lower-level / short time-scale}: a RNN-based language model $f$ encoding representations in the activations of a hidden state,
\item \textit{Middle-level / medium time-scale}: a meta-learner $g$ updating the set of weights of the language model to store medium-term representations, and
\item \textit{Higher-level / long time-scale}: a static long-term memory of the dynamic of the RNN-based language model (see below).
\end{enumerate}

\begin{figure*}
\centering
\includegraphics[width=1\linewidth]{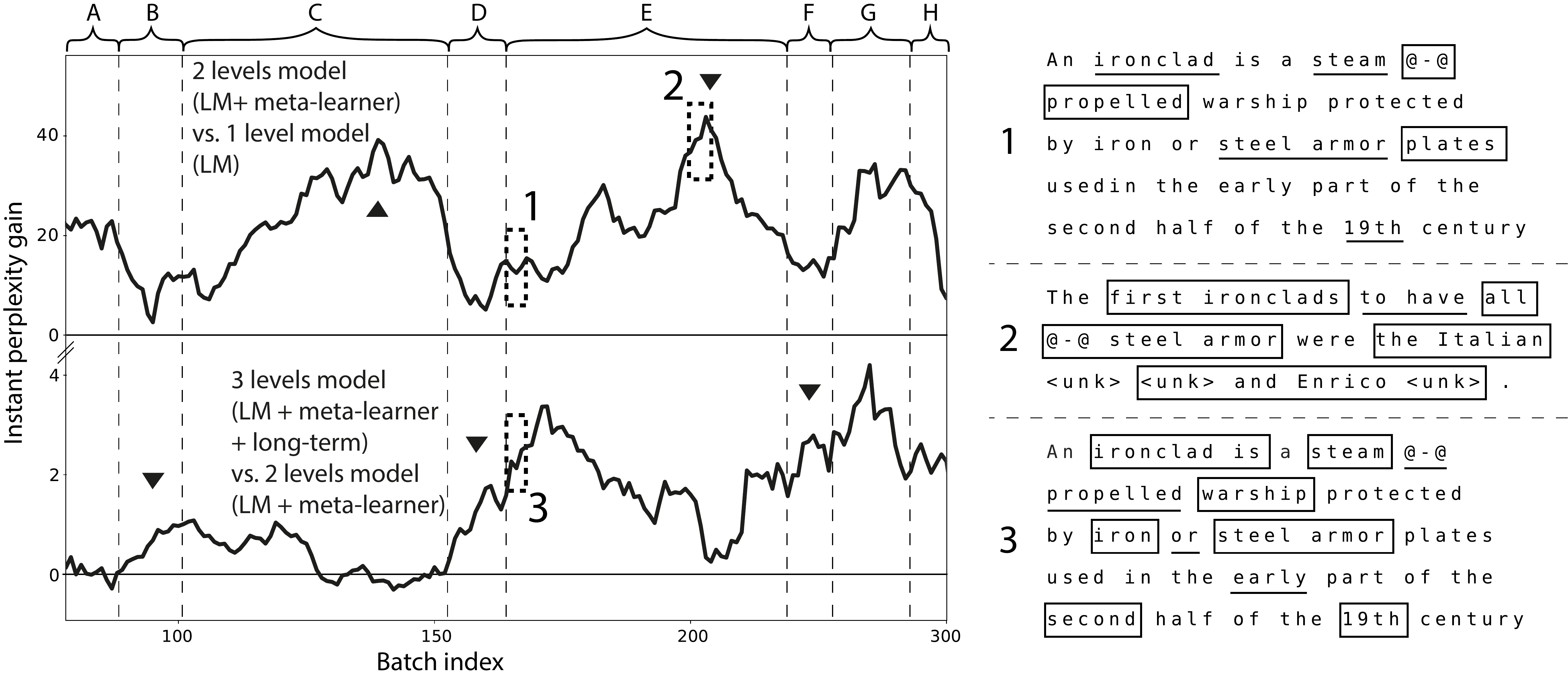}
\caption{Medium and long-term memory effects on a sample of Wikitext-2 test set with a sequence of Wikipedia articles (letters $A-H$). (Left) Instantaneous perplexity gain: difference in batch perplexity between models. Higher values means the first model has locally a lower perplexity than the second model. (Top curve) Comparing a two-levels model (LM + meta-learner) with a one-level model (LM). (Bottom curve) Comparing a three-levels model (LM + meta-learner + long-term memory) with a two-levels model. (Right) Token loss difference on three batch samples indicated on the left curves. A squared (resp. underlined) word means the first model has a lower (resp. higher) loss on that word than the second model. We emphasize only words associated with a significant difference in loss by setting a threshold at 10 percent of the maximum absolute loss of each sample.}
\label{fig:fig2}
\end{figure*}

The meta-learner $g$ is trained to update the lower-level network $f$ by computing $f_t, i_t, z_t=g_\phi(\theta_{t-1}, \mathcal{L}_t^{LM}, \nabla_{\theta_{t-1}}\mathcal{L}_t^{LM}, \theta_{0})$ and updating the set of weights as
\begin{equation}
\theta_{t} = f_t\odot \theta_{t-1} +i_t\odot\nabla_{\theta_{t-1}}\mathcal{L}_t^{LM} + z_t\odot \theta_{0}
\end{equation}
This hierarchical network could be seen as an analog of the hierarchical recurrent neural networks \citep{chung_hierarchical_2016} where the gates $f_t$, $i_t$ and $z_t$ can be seen as controlling a set of COPY, FLUSH and UPDATE operations:
\begin{enumerate}
\item COPY ($f_t$): part of the state copied from the previous state $\theta_{t-1}$,
\item UPDATE ($i_t$): part of the state updated by the loss gradients on the previous batch, and
\item FLUSH ($z_t$): part of the state reset from a static long term memory $\theta_{0}$.
\end{enumerate}
One difference with the work of \cite{chung_hierarchical_2016} is that the memory was confined to the hidden in the later while the memory of our hierarchical network is split between the weights of the lower-level network and its hidden-state.

The meta-learner can be a feed-forward or a RNN network. In our experiments, simple linear feed-forward networks lead to the lower perplexities, probably because it was easier to regularize and optimize. The meta-learner implements coordinate-sharing as described in \cite{andrychowicz_learning_2016,ravi_optimization_2016} and takes as input the loss $\mathcal{L}_t^{LM}$ and loss-gradients $\nabla_{\theta_{t-1}}\mathcal{L}_t^{LM}$ over a previous batch $B_i$ (a sequence of $M$ tokens $w_{0},\dots, w_{M}$ as illustrated on figure \ref{fig:fig1}). The size $M$ of the batch adjusts the trade-off between the noise of the loss/gradients and updating frequency of the medium-term memory, smaller batches leading to faster updates with higher noise.

\subsection{Continual learning}
\label{sec:continuous}
The interaction between the meta-learner and the language model implements a form of continual-learning and the language model thus faces a phenomenon known as catastrophic forgetting \citep{french_catastrophic_1999}. In our case, this correspond to the lower-level network over-specializing to the lexical field of a particular topic after several updates of the meta-learner (e.g. while processing a long article on a specific topic).

To mitigate this effect we use a higher-level static memory initialized using "elastic weight consolidation" (EWC) introduced by Kirkpatrick et al. (\citeyear{kirkpatrick_overcoming_2017}) to reduce forgetting in multi-task reinforcement learning.

Casting our task in the EWC framework, we define a task A which is the language modeling task (prediction of next token) when no context is stored in the weights of the lower-level network. The solution of task A is a set of weights toward which the model could advantageously come back when the context stored in the weights become irrelevant (for example when switching between paragraphs on different topics). To obtain a set of weights for task A, we train the lower-level network (RNN) alone on the training dataset and obtain a set of weights that would perform well on average, i.e. when no specific context has been provided by a context-dependent weight update performed by the meta-learner.

We then define a task B which is a language modeling task when a context has been stored in the weights of the lower-level network by an update of the meta-learner. The aim of EWC is to learn task B while retaining some performance on task A.

Empirical results suggest that many weights configurations result in similar performances \citep{sussmann_uniqueness_1992} and there is thus likely a solution for task B close to a solution for task A. The idea behind EWC is to learn task B while protecting the performance in task A by constraining the parameters to stay around the solution found for task A.

This constraint is implemented as a quadratic penalty, similarly to spring anchoring the parameters, hence the name elastic. The stiffness of the springs should be greater for parameters that most affect performance in task A. We can formally write this constrain by using Bayes rule to express the conditional log probability of the parameters when the training dataset $\mathcal{D}$ is split between the training dataset for task A ($\mathcal{D}_A$) and the training dataset for task B ($\mathcal{D}_B$):
\begin{equation}
\log p(\theta\mid\mathcal{D}) = \log p(\mathcal{D}_B\mid\theta) + \log p(\theta\mid\mathcal{D}_A) - \log p(\mathcal{D}_B)
\end{equation}
The true posterior probability on task A $p(\theta\mid\mathcal{D}_A)$ is intractable, so we approximate the posterior as a Gaussian distribution with mean given by the parameters and a diagonal precision given by the diagonal of the Fisher information matrix F which is equivalent to the second derivative of the loss near a minimum and can be computed from first-order derivatives alone.

\section{Related work}
Several works have been devoted to dynamically updating the weights of neural networks during inference. A few recent architectures are the Fast-Weights of \citet{ba_using_2016}, the Hypernetworks of \citet{ha_hypernetworks_2016} and the Nested LSTM of \citet{moniz_nested_2018}. The weights update rules of theses models use as inputs one or several of (i) a previous hidden state of a RNN network or higher level network and/or (ii) the current or previous inputs to the network. However, these models do not use the predictions of the network on the previous tokens (i.e. the loss and gradient of the loss of the model) as in the present work.
The architecture that is most related to the present work is the study on dynamical evaluation of \citet{krause_dynamic_2017} in which a loss function similar to the loss function of the present work is obtained empirically and optimized using a large hyper-parameter search on the parameters of the SGD-like rule.

\section{Experiments}
\label{sec:3}
\subsection{Architecture and hyper-parameters}
\label{sec:training}
As mentioned in \ref{sec:continuous}, a set of pre-trained weights of the RNN language model is first obtained by training the lower-level network $f$ and computing the diagonal of the Fisher matrix around the final weights.

Then, the meta-learner $g$ is trained in an online meta-learning fashion on the validation dataset (alternatively, a sub-set of the training dataset could be used). A training sequence $S$ is split in a sequence of mini-batches $B_i$, each batch $B_i$ containing $M$ inputs tokens ($w_{i\times M},\dots, w_{i\times M+M}$) and $M$ associated targets ($w_{i\times M+1},\dots, w_{i\times M+M+1}$). In our experiments we varied $M$ between 5 and 20.

The meta-learner is trained as described in \citep{andrychowicz_learning_2016,li_learning_2016} by minimizing the sum over the sequence of LM losses: $\mathcal{L}_{meta}=\sum_{i>0} \mathcal{L}_i^{LM}$. The meta-learner is trained by truncated back-propagation through time and is unrolled over at least 40 steps as the reward from the medium-term memory is relatively sparse \cite{li_learning_2016}.

To be able to unroll the model over a sufficient number of steps while using a state-of-the-art language model with over than 30 millions parameters, we use a memory-efficient version of back propagation through time based on gradient checkpointing as described by Grusly et al. (\citeyear{gruslys_memory-efficient_2016}).

\subsection{Experiments}
We performed a series of experiments on the Wikitext-2 dataset \cite{merity_pointer_2016} using an AWD-LSTM language model \citep{merity_regularizing_2017} and a feed-forward and RNN meta-learner.

The test perplexity are similar to perplexities obtained using dynamical evaluation \citep{krause_dynamic_2017}, reaching $46.9$ with a linear feed-forward meta-learner when starting from a one-level language model with test perplexity of $64.8$.

In our experiments, the perplexity could not be improved by using a RNN meta-learner or a deeper meta-learner. We hypothesis that this may be caused by several reasons. First, storing a hidden state in the meta-learner might be less important in an online meta-learning setup than it is in a standard meta-learning setup \citep{andrychowicz_learning_2016} as the target distribution of the weights is non-stationary. Second, the size of the hidden state cannot be increased significantly without reducing the number of steps along which the meta-learner is unrolled during meta-training which may be detrimental. 

Some quantitative experiments are shown on Figure \ref{fig:fig2} using a linear feed-forward network to illustrate the effect of the various layers in the hierarchical model. The curves shows differences in batch perplexity between model variants.

The top curve compares a one-level model (language model) with a two-levels model (language model + meta-learner). The meta-learner is able to learn medium-term representations to progressively reduce perplexity along articles (see e.g. articles C and E). Right sample 1 (resp. 2) details sentences at the begging (resp. middle) of article E related to a warship called "Ironclad". The addition of the meta-learner reduces the loss on a number of expression related to the warship like "ironclad" or "steel armor".

Bottom curve compares a three-levels model (language model + meta-learner + long-term memory) with the two-levels model. The local loss is reduced at topics changes and beginning of new topics (see e.g. articles B, D and F). The right sample 3 can be contrasted with sample 1 to illustrate how the hierarchical model is able to better recover a good parameter space following a change in topic.

\bibliography{Zotero}

\begin{thebibliography}{34}
\expandafter\ifx\csname natexlab\endcsname\relax\def\natexlab#1{#1}\fi

\bibitem[{Abbott and Regehr(2004)}]{abbott_synaptic_2004}
L.~F. Abbott and Wade~G. Regehr. 2004.
\newblock \href {https://doi.org/10.1038/nature03010} {Synaptic computation}.
\newblock \emph{Nature}, 431(7010):796--803.

\bibitem[{Andrychowicz et~al.(2016)Andrychowicz, Denil, Gomez, Hoffman, Pfau,
  Schaul, Shillingford, and de~Freitas}]{andrychowicz_learning_2016}
Marcin Andrychowicz, Misha Denil, Sergio Gomez, Matthew~W. Hoffman, David Pfau,
  Tom Schaul, Brendan Shillingford, and Nando de~Freitas. 2016.
\newblock \href {http://arxiv.org/abs/1606.04474} {Learning to learn by
  gradient descent by gradient descent}.
\newblock \emph{arXiv:1606.04474 [cs]}.
\newblock ArXiv: 1606.04474.

\bibitem[{Arisoy et~al.(2012)Arisoy, Sainath, Kingsbury, and
  Ramabhadran}]{arisoy_deep_2012}
Ebru Arisoy, Tara~N. Sainath, Brian Kingsbury, and Bhuvana Ramabhadran. 2012.
\newblock Deep neural network language models.
\newblock In \emph{Proceedings of the {NAACL}-{HLT} 2012 {Workshop}: {Will}
  {We} {Ever} {Really} {Replace} the {N}-gram {Model}? {On} the {Future} of
  {Language} {Modeling} for {HLT}}, pages 20--28. Association for Computational
  Linguistics.

\bibitem[{Ba et~al.(2016)Ba, Hinton, Mnih, Leibo, and Ionescu}]{ba_using_2016}
Jimmy Ba, Geoffrey Hinton, Volodymyr Mnih, Joel~Z. Leibo, and Catalin Ionescu.
  2016.
\newblock \href {http://arxiv.org/abs/1610.06258} {Using {Fast} {Weights} to
  {Attend} to the {Recent} {Past}}.
\newblock \emph{arXiv:1610.06258 [cs, stat]}.
\newblock ArXiv: 1610.06258.

\bibitem[{Bahdanau et~al.(2014)Bahdanau, Cho, and
  Bengio}]{bahdanau_neural_2014}
Dzmitry Bahdanau, Kyunghyun Cho, and Yoshua Bengio. 2014.
\newblock \href {http://arxiv.org/abs/1409.0473} {Neural machine translation by
  jointly learning to align and translate}.
\newblock \emph{arXiv preprint arXiv:1409.0473}.

\bibitem[{Barak and Tsodyks(2007)}]{barak_persistent_2007}
Omri Barak and Misha Tsodyks. 2007.
\newblock \href {https://doi.org/10.1371/journal.pcbi.0030035} {Persistent
  activity in neural networks with dynamic synapses}.
\newblock \emph{PLoS computational biology}, 3(2):e35.

\bibitem[{Cho et~al.(2014)Cho, van Merrienboer, Gulcehre, Bahdanau, Bougares,
  Schwenk, and Bengio}]{cho_learning_2014}
Kyunghyun Cho, Bart van Merrienboer, Caglar Gulcehre, Dzmitry Bahdanau, Fethi
  Bougares, Holger Schwenk, and Yoshua Bengio. 2014.
\newblock \href {http://arxiv.org/abs/1406.1078} {Learning {Phrase}
  {Representations} using {RNN} {Encoder}-{Decoder} for {Statistical} {Machine}
  {Translation}}.
\newblock \emph{arXiv:1406.1078 [cs, stat]}.
\newblock ArXiv: 1406.1078.

\bibitem[{Chung et~al.(2016)Chung, Ahn, and Bengio}]{chung_hierarchical_2016}
Junyoung Chung, Sungjin Ahn, and Yoshua Bengio. 2016.
\newblock \href {http://arxiv.org/abs/1609.01704} {Hierarchical {Multiscale}
  {Recurrent} {Neural} {Networks}}.
\newblock \emph{arXiv:1609.01704 [cs]}.
\newblock ArXiv: 1609.01704.

\bibitem[{Daniluk et~al.(2016)Daniluk, Rocktäschel, Welbl, and
  Riedel}]{daniluk_frustratingly_2016}
Michał Daniluk, Tim Rocktäschel, Johannes Welbl, and Sebastian Riedel. 2016.
\newblock \href {https://openreview.net/forum?id=ByIAPUcee} {Frustratingly
  {Short} {Attention} {Spans} in {Neural} {Language} {Modeling}}.

\bibitem[{El~Hihi and Bengio(1995)}]{el_hihi_hierarchical_1995}
Salah El~Hihi and Yoshua Bengio. 1995.
\newblock \href {http://dl.acm.org/citation.cfm?id=2998828.2998898}
  {Hierarchical {Recurrent} {Neural} {Networks} for {Long}-term
  {Dependencies}}.
\newblock In \emph{Proceedings of the 8th {International} {Conference} on
  {Neural} {Information} {Processing} {Systems}}, {NIPS}'95, pages 493--499,
  Cambridge, MA, USA. MIT Press.

\bibitem[{French(1999)}]{french_catastrophic_1999}
Robert~M. French. 1999.
\newblock \href {https://doi.org/10.1016/S1364-6613(99)01294-2} {Catastrophic
  forgetting in connectionist networks}.
\newblock \emph{Trends in Cognitive Sciences}, 3(4):128--135.

\bibitem[{Graves et~al.(2014)Graves, Wayne, and Danihelka}]{graves_neural_2014}
Alex Graves, Greg Wayne, and Ivo Danihelka. 2014.
\newblock \href {http://arxiv.org/abs/1410.5401} {Neural turing machines}.
\newblock \emph{arXiv preprint arXiv:1410.5401}.

\bibitem[{Gruslys et~al.(2016)Gruslys, Munos, Danihelka, Lanctot, and
  Graves}]{gruslys_memory-efficient_2016}
Audrūnas Gruslys, Remi Munos, Ivo Danihelka, Marc Lanctot, and Alex Graves.
  2016.
\newblock \href {http://arxiv.org/abs/1606.03401} {Memory-{Efficient}
  {Backpropagation} {Through} {Time}}.
\newblock \emph{arXiv:1606.03401 [cs]}.
\newblock ArXiv: 1606.03401.

\bibitem[{Ha et~al.(2016)Ha, Dai, and Le}]{ha_hypernetworks_2016}
David Ha, Andrew Dai, and Quoc~V. Le. 2016.
\newblock \href {http://arxiv.org/abs/1609.09106} {{HyperNetworks}}.
\newblock \emph{arXiv:1609.09106 [cs]}.
\newblock ArXiv: 1609.09106.

\bibitem[{Hochreiter and Schmidhuber(1997)}]{hochreiter_long_1997}
Sepp Hochreiter and Jürgen Schmidhuber. 1997.
\newblock \href {https://doi.org/10.1162/neco.1997.9.8.1735} {Long
  {Short}-{Term} {Memory}}.
\newblock \emph{Neural Comput.}, 9(8):1735--1780.

\bibitem[{Howard and Ruder(2018)}]{howard_fine-tuned_2018}
Jeremy Howard and Sebastian Ruder. 2018.
\newblock \href {http://arxiv.org/abs/1801.06146} {Fine-tuned {Language}
  {Models} for {Text} {Classification}}.
\newblock \emph{arXiv:1801.06146 [cs, stat]}.
\newblock ArXiv: 1801.06146.

\bibitem[{Kirkpatrick et~al.(2017)Kirkpatrick, Pascanu, Rabinowitz, Veness,
  Desjardins, Rusu, Milan, Quan, Ramalho, Grabska-Barwinska, Hassabis, Clopath,
  Kumaran, and Hadsell}]{kirkpatrick_overcoming_2017}
James Kirkpatrick, Razvan Pascanu, Neil Rabinowitz, Joel Veness, Guillaume
  Desjardins, Andrei~A. Rusu, Kieran Milan, John Quan, Tiago Ramalho, Agnieszka
  Grabska-Barwinska, Demis Hassabis, Claudia Clopath, Dharshan Kumaran, and
  Raia Hadsell. 2017.
\newblock \href {https://doi.org/10.1073/pnas.1611835114} {Overcoming
  catastrophic forgetting in neural networks}.
\newblock \emph{Proceedings of the National Academy of Sciences},
  114(13):3521--3526.

\bibitem[{Koutník et~al.(2014)Koutník, Greff, Gomez, and
  Schmidhuber}]{koutnik_clockwork_2014}
Jan Koutník, Klaus Greff, Faustino Gomez, and Jürgen Schmidhuber. 2014.
\newblock \href {http://dl.acm.org/citation.cfm?id=3044805.3045100} {A
  {Clockwork} {RNN}}.
\newblock In \emph{Proceedings of the 31st {International} {Conference} on
  {International} {Conference} on {Machine} {Learning} - {Volume} 32},
  {ICML}'14, pages II--1863--II--1871, Beijing, China. JMLR.org.

\bibitem[{Krause et~al.(2017)Krause, Kahembwe, Murray, and
  Renals}]{krause_dynamic_2017}
Ben Krause, Emmanuel Kahembwe, Iain Murray, and Steve Renals. 2017.
\newblock \href {http://arxiv.org/abs/1709.07432} {Dynamic {Evaluation} of
  {Neural} {Sequence} {Models}}.
\newblock \emph{arXiv:1709.07432 [cs]}.
\newblock ArXiv: 1709.07432.

\bibitem[{Li and Malik(2016)}]{li_learning_2016}
Ke~Li and Jitendra Malik. 2016.
\newblock \href {http://arxiv.org/abs/1606.01885} {Learning to {Optimize}}.
\newblock \emph{arXiv:1606.01885 [cs, math, stat]}.
\newblock ArXiv: 1606.01885.

\bibitem[{Merity et~al.(2017)Merity, Keskar, and
  Socher}]{merity_regularizing_2017}
Stephen Merity, Nitish~Shirish Keskar, and Richard Socher. 2017.
\newblock \href {http://arxiv.org/abs/1708.02182} {Regularizing and
  {Optimizing} {LSTM} {Language} {Models}}.
\newblock \emph{arXiv:1708.02182 [cs]}.
\newblock ArXiv: 1708.02182.

\bibitem[{Merity et~al.(2016)Merity, Xiong, Bradbury, and
  Socher}]{merity_pointer_2016}
Stephen Merity, Caiming Xiong, James Bradbury, and Richard Socher. 2016.
\newblock \href {http://arxiv.org/abs/1609.07843} {Pointer {Sentinel} {Mixture}
  {Models}}.
\newblock \emph{arXiv:1609.07843 [cs]}.
\newblock ArXiv: 1609.07843.

\bibitem[{Mikolov et~al.(2010)Mikolov, Karafiát, Burget, Cernocký, and
  Khudanpur}]{mikolov_recurrent_2010}
Tomas Mikolov, Martin Karafiát, Lukas Burget, Jan Cernocký, and Sanjeev
  Khudanpur. 2010.
\newblock \emph{Recurrent neural network based language model}, volume~2.

\bibitem[{Moniz and Krueger(2018)}]{moniz_nested_2018}
Joel Ruben~Antony Moniz and David Krueger. 2018.
\newblock \href {http://arxiv.org/abs/1801.10308} {Nested {LSTMs}}.
\newblock \emph{arXiv:1801.10308 [cs]}.
\newblock ArXiv: 1801.10308.

\bibitem[{Nallapati et~al.(2016)Nallapati, Zhou, Gulcehre, and
  Xiang}]{nallapati_abstractive_2016}
Ramesh Nallapati, Bowen Zhou, Caglar Gulcehre, and Bing Xiang. 2016.
\newblock Abstractive text summarization using sequence-to-sequence rnns and
  beyond.
\newblock \emph{arXiv preprint arXiv:1602.06023}.

\bibitem[{Peters et~al.(2018)Peters, Neumann, Iyyer, Gardner, Clark, Lee, and
  Zettlemoyer}]{peters_deep_2018}
Matthew~E. Peters, Mark Neumann, Mohit Iyyer, Matt Gardner, Christopher Clark,
  Kenton Lee, and Luke Zettlemoyer. 2018.
\newblock \href {http://arxiv.org/abs/1802.05365} {Deep contextualized word
  representations}.
\newblock \emph{arXiv:1802.05365 [cs]}.
\newblock ArXiv: 1802.05365.

\bibitem[{Radford et~al.(2017)Radford, Jozefowicz, and
  Sutskever}]{radford_learning_2017}
Alec Radford, Rafal Jozefowicz, and Ilya Sutskever. 2017.
\newblock \href {http://arxiv.org/abs/1704.01444} {Learning to {Generate}
  {Reviews} and {Discovering} {Sentiment}}.
\newblock \emph{arXiv:1704.01444 [cs]}.
\newblock ArXiv: 1704.01444.

\bibitem[{Ravi and Larochelle(2016)}]{ravi_optimization_2016}
Sachin Ravi and Hugo Larochelle. 2016.
\newblock \href {https://openreview.net/forum?id=rJY0-Kcll} {Optimization as a
  {Model} for {Few}-{Shot} {Learning}}.

\bibitem[{Rush et~al.(2015)Rush, Chopra, and Weston}]{rush_neural_2015}
Alexander~M. Rush, Sumit Chopra, and Jason Weston. 2015.
\newblock A neural attention model for abstractive sentence summarization.
\newblock \emph{arXiv preprint arXiv:1509.00685}.

\bibitem[{Schmidhuber(1992)}]{schmidhuber_learning_1992}
J.~Schmidhuber. 1992.
\newblock \href {https://doi.org/10.1162/neco.1992.4.2.234} {Learning
  {Complex}, {Extended} {Sequences} {Using} the {Principle} of {History}
  {Compression}}.
\newblock \emph{Neural Computation}, 4(2):234--242.

\bibitem[{Schwenk(2012)}]{schwenk_continuous_2012}
Holger Schwenk. 2012.
\newblock Continuous space translation models for phrase-based statistical
  machine translation.
\newblock \emph{Proceedings of COLING 2012: Posters}, pages 1071--1080.

\bibitem[{Sussmann(1992)}]{sussmann_uniqueness_1992}
Héctor~J. Sussmann. 1992.
\newblock \href {https://doi.org/10.1016/S0893-6080(05)80037-1} {Uniqueness of
  the weights for minimal feedforward nets with a given input-output map}.
\newblock \emph{Neural Networks}, 5(4):589--593.

\bibitem[{Tran et~al.(2016)Tran, Bisazza, and Monz}]{tran_recurrent_2016}
Ke~Tran, Arianna Bisazza, and Christof Monz. 2016.
\newblock \href {http://arxiv.org/abs/1601.01272} {Recurrent {Memory}
  {Networks} for {Language} {Modeling}}.
\newblock \emph{arXiv:1601.01272 [cs]}.
\newblock ArXiv: 1601.01272.

\bibitem[{Tsodyks et~al.(1998)Tsodyks, Pawelzik, and
  Markram}]{tsodyks_neural_1998}
Misha Tsodyks, Klaus Pawelzik, and Henry Markram. 1998.
\newblock \emph{Neural {Networks} with {Dynamic} {Synapses}}.

\end{thebibliography}
\bibliographystyle{acl_natbib}

\end{document}